\documentclass[final]{cvpr}

\usepackage{times}
\usepackage{epsfig}
\usepackage{graphicx}
\usepackage{amsmath}
\usepackage{amssymb}
\usepackage{pifont}
\usepackage{dsfont}
\usepackage{array}
\usepackage{booktabs}
\usepackage{dblfloatfix}
\usepackage{enumitem}
\usepackage{multicol}
\usepackage{multirow}
\usepackage{adjustbox}
\usepackage{subcaption}
\newcommand{\bigcell}[2]{\begin{tabular}{@{}#1@{}}#2\end{tabular}}
\newcolumntype{P}[1]{>{\centering\arraybackslash}p{#1}}
\newcolumntype{M}[1]{>{\centering\arraybackslash}m{#1}}
\newcommand{\cmark}{\text{\ding{51}}}

\setlist[itemize]{leftmargin=4.mm}
\newcommand*\samethanks[1][\value{footnote}]{\footnotemark[#1]}

\usepackage[pagebackref=true,breaklinks=true,colorlinks,bookmarks=false]{hyperref}

\begin{document}

\title{Pose Recognition with Cascade Transformers}

\author{Ke Li\thanks{~indicates equal contribution.}$~~^1$, Shijie Wang\samethanks$~~^2$, Xiang Zhang\samethanks$~~^2$, Yifan Xu$^3$, Weijian Xu$^3$, Zhuowen Tu$^3$\\
$^1$University of Chinese Academy of Sciences, Beijing, China\\
$^2$Tsinghua University, Beijing, China\\
$^3$University of California San Diego, San Diego, USA\\
{\tt\small \{keliictcas, wang98thu, zx1239856\}@gmail.com, \{yix081, wex041, ztu\}@ucsd.edu}}

\vspace{-0.5cm}
\maketitle
\begin{abstract}
In this paper, we present a regression-based pose recognition method using cascade Transformers. One way to categorize the existing approaches in this domain is to separate them into 1). heatmap-based and 2). regression-based. In general, heatmap-based methods achieve higher accuracy but are subject to various heuristic designs (not end-to-end mostly), whereas regression-based approaches attain relatively lower accuracy but they have less intermediate non-differentiable steps. Here we utilize the encoder-decoder structure in Transformers to perform regression-based person and keypoint detection that is general-purpose and requires less heuristic design compared with the existing approaches. We demonstrate the keypoint hypothesis (query) refinement process across different self-attention layers to reveal the recursive self-attention mechanism in Transformers. In the experiments, we report competitive results for pose recognition when compared with the competing regression-based methods.
\let\thefootnote\relax\footnotetext{Code: \url{https://github.com/mlpc-ucsd/PRTR}.}
\let\thefootnote\relax\footnotetext{Work performed during internships of K. Li, S.Wang, and X. Zhang with UC San Diego.}

\end{abstract}
\vspace{-3mm}
\section{Introduction}

We tackle the 2D human pose recognition problem \cite{lin2014microsoft,andriluka20142d,toshev2014deeppose,newell2016stacked} where keypoints (\eg head, shoulders, knees, \etc) for multiple people in an RGB image are to be detected and localized. This is an important problem in computer vision that can be adopted in a variety of downstream tasks including tracking, security, animation, human-computer interaction, computer games, and robotics.

There has been a steady progress in 2D human pose recognition \cite{andriluka20142d,toshev2014deeppose,wei2016convolutional,newell2016stacked,kreiss2019pifpaf,cao2017realtime,papandreou2017towards,sun2018integral,papandreou2018personlab,cheng2020higherhrnet,chen2018cascaded,sun2019deep,zhou2019objects,nie2019single} with systems becoming increasingly practical without a strong constraint (\eg present multiple people of varying size). However, pose recognition is a challenging problem that remains unsolved. The difficulty lies in various aspects such as large pose/shape variation, inter-person and self occlusion, large appearance variation, and background clutter.

\begin{figure}[!t]
\centering
\includegraphics[width=\linewidth,trim=0 0.3cm 1.2cm 0,clip]{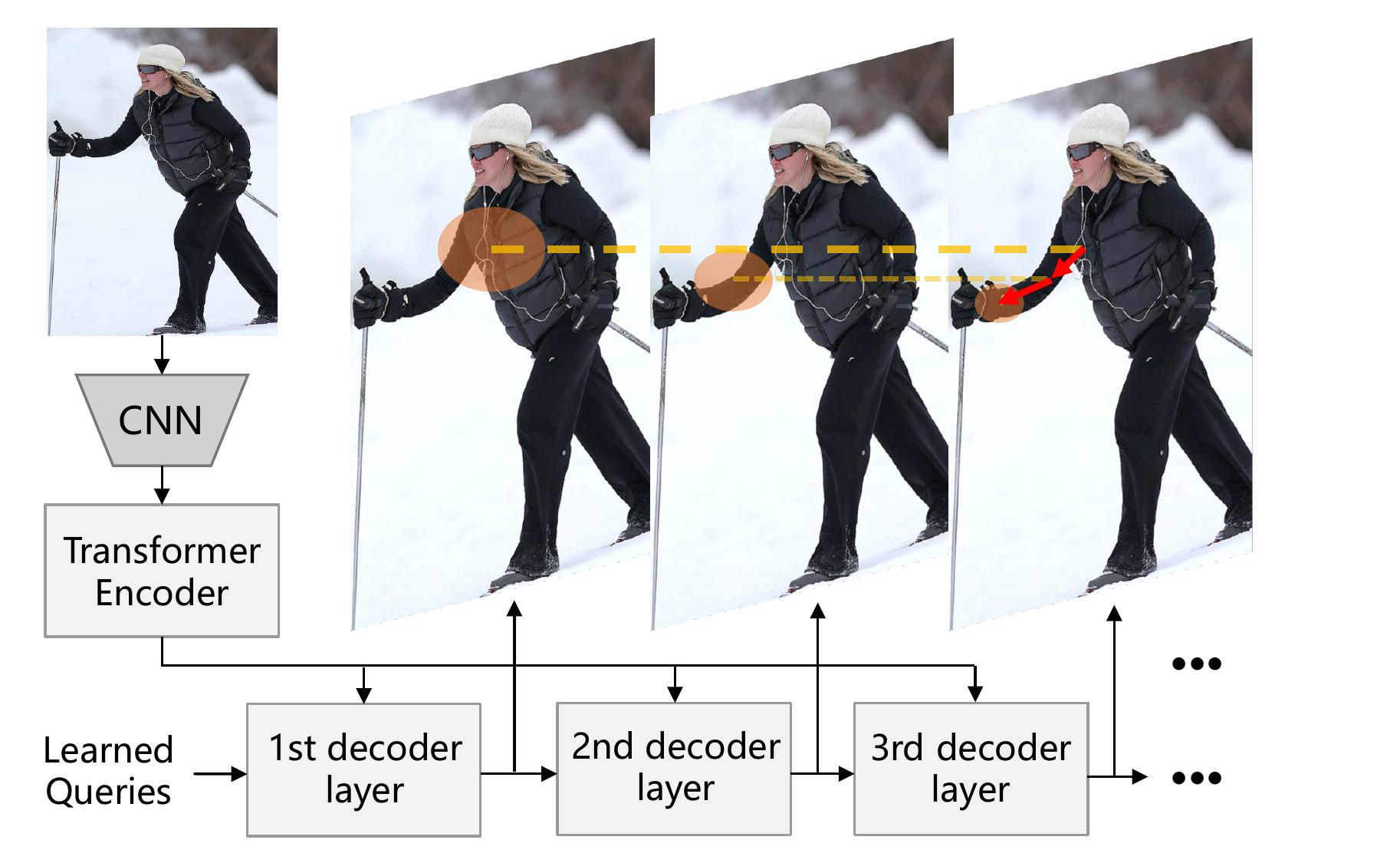}
\vspace{-2mm}
\caption{Illustration of the gradual refinement for the keypoints across different Transformer decoder layers. Through the decoding process, PRTR predicts keypoints with increasing confidence and decreasing spatial deviation to ground truth, transforming image-ignorant queries to final predictions.}  
\vspace{-5mm}
\label{teaser}
\end{figure}

For multiple people in an input image \cite{lin2014microsoft}, the task of pose recognition is to localize the human keypoints (17 in the experiments) for the individual persons. This can be achieved by a two-stage process in which individual persons are detected first, followed by keypoint detection from the detected image region/patch; this is called a top-down process \cite{sun2019deep}. An alternative strategy is called a bottom-up process where human keypoints are detected directly from the image without an explicit object detection stage \cite{cheng2020higherhrnet}. A discussion about the top-down and bottom-up approaches can be found in \cite{cheng2020higherhrnet}.

Another way to divide the existing literature in pose recognition is based on the choice of using heatmap or regression. Heatmap-based approaches \cite{xiao2018simple,sun2019deep} perform dense keypoint detection followed by subsequent processes for clustering and grouping; they deliver strong performance but are also subject to many heuristic designs that are mostly not end-to-end learnable. Regression based methods \cite{sun2018integral,zhou2019objects,wei2020point} perform regression for the keypoints directly which have less intermediate stages and specifications. Regression-based methods typically perform worse than heatmap-based ones, but can be made end-to-end and readily integrated with the other downstream tasks. Reasons for the existence of both heatmap-based and regression-based methods are present. Heatmap-based methods are adopted when the accuracy is the priority whereas regression-based approaches can be considered as a convenient plug-and-play module. 

Generally, heatmap-based methods adopt handcrafted or heuristic pre/post-processing to encode ground truth to heatmaps and decode heatmaps to predict keypoints. These methods introduce design challenges and biases, making them sub-optimal. They are hard to update and adapt as well. In detail, SimpleBaseline \cite{xiao2018simple} and HRNet \cite{sun2019deep} adopt the standard coordinate decoding method designed empirically according to model performance in \cite{newell2016stacked}, refining the coordinates 0.25 time from the maximum activation to the second maximum empirically in the heatmap. DARK \cite{zhang2020distribution} presents Taylor-expansion based coordinate decoding and unbiased sub-pixel centered coordinate encoding. UDP \cite{Huang_2020_CVPR} even discovered a considerable accuracy decrease when using one-pixel flip shift in heatmap-based paradigms. For general-purpose regression methods, we aim at removing unnecessary designs by making the training objective and target output direct and transparent. Coordinates should be output directly and the loss be calculated with predictions and ground truth coordinates straightforward.

\begin{figure*}[!t]
\centering
\includegraphics[width=1\linewidth]{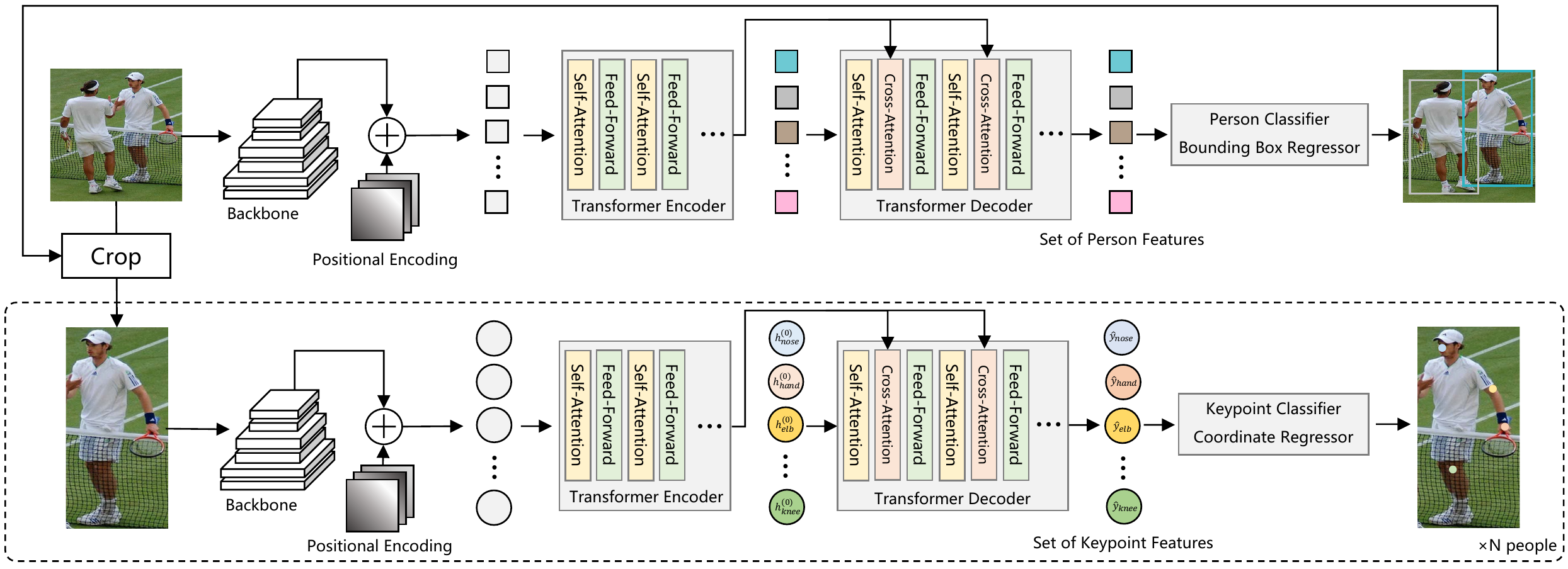}
\vspace{-4mm}
\caption{\small The architecture of Pose Recognition with TRansformer (\textbf{PRTR}), \textbf{two-stage variant}. First, using whole-picture image feature and absolute positional encoding, a person-detection Transformer detects people in the image with a set of learned person queries. After filtering background queries, we crop the original image with predicted boxes. Cropped images are fed into a keypoint-detection Transformer, together with positional encoding relative to corresponding bounding boxes. Finally, we read out $J$ keypoints from a larger set of keypoint queries by Hungarian algorithm. The keypoint-detection Transformer processes all the non-background keypoint proposals in a vectorized way. $h_{\_}^{(0)}$ denotes hypotheses (queries), the feature vectors to be refined to final predictions, $\hat{y}_{\_}$, through Transformer decoder.}\label{fig:model1}
\vspace{-3mm}
\end{figure*}

Bearing this in mind, we present a top-down regression-based 2D human pose recognition method using cascade Transformers consisting of a person detection Transformer and a keypoint detection Transformer. Two alternatives have been developed, one being a two-stage process (shown in Figure~\ref{fig:model1}) with the two Transformers learned sequentially and the other being a sequential process (shown in Figure~\ref{fig:model2}) with the two transfomers learned jointly in an end-to-end fashion. We name our method Pose Regression TRansformers (PRTR). We apply multi-scale features in the keypoint detection Transformer. Visualization for the keypoint queries across different attention layers in the decoder is given to illustrate the internal detection process. PRTR is a general-purpose approach for keypoint regression and we show competitive results in pose recognition when compared with the existing regression-based methods in the literature. The contributions of our work include:
\begin{itemize}
\setlength\itemsep{0mm}
    \item We propose a regression-based human pose recognition method by building cascade Transformers, based on a general-purpose object detector, end-to-end object detection Transformer (DETR) \cite{carion2020end}. Our method, named pose recognition Transformer (PRTR), enjoys the tokenized representation in Transformers with layers of self-attention to capture the joint spatial and appearance modeling for the keypoints. 
    \item Two types of cascade Transformers have been developed: 1). a two-stage one with the second Transformer taking image patches detected from the first Transformer, as shown in Figure~\ref{fig:model1}; and 2). a sequential one using spatial Transformer network (STN) \cite{jaderberg2015spatial} to create an end-to-end framework, shown in Figure~\ref{fig:model2}.
    \item We visualize the distribution of keypoint queries in various aspects to unfold the internal process of the Transformer for the gradual refinement of the detection. 
\end{itemize}
On the COCO 2D human pose recognition dataset \cite{lin2014microsoft}, competitive results have been observed when compared with the regression-based methods.

\section{Related Work}

Given an image $I$, the goal of pose recognition is to predict a possibly empty set of persons, $\{P_i\}_{i=1}^N$, where $N$ is the number of persons in the image. For each person, we need to predict its bounding box position, $b_i$, as well as its skeleton coordinates, $s_i = \{(x_j, y_j)\}_{j=1}^J$, where $J$ is the number of joints pre-defined in each dataset.

We discuss related work from several aspects.
The field of human pose regression has witnessed a continuing progress \cite{andriluka20142d,toshev2014deeppose,wei2016convolutional,newell2016stacked,kreiss2019pifpaf,cao2017realtime,papandreou2017towards,sun2018integral,papandreou2018personlab,cheng2020higherhrnet,chen2018cascaded,sun2019deep,zhou2019objects,nie2019single}, in particular with the advancing of the deep learning technologies \cite{krizhevsky2012imagenet,goodfellow2016deep,he2016deep}. One notable development in pose recognition is the creation of the HRNet family model \cite{sun2019deep,cheng2020higherhrnet} which is itself about a new convolutional neural network (CNN) architecture targeting the modeling of high-resolution feature responses. HRNet \cite{sun2019deep} has shown its particular advantage in advancing the state-of-the-art for 2D human pose recognition/estimation.

\textbf{Heatmap-based} approaches include \cite{cao2017realtime,he2017mask,papandreou2017towards,newell2017associative,kreiss2019pifpaf,papandreou2018personlab,cheng2020higherhrnet,chen2018cascaded,xiao2018simple,sun2019deep,zhang2020distribution,Yang_2017_ICCV,Tang_2018_ECCV} where various techniques have been developed to perform multi-class keypoint classification. The classifiers produce dense heatmaps (classification map), followed by clustering and grouping processes. On one hand, heatmap-based methods leverage fine-grained detection for the keypoints by densely scanning all the pixels; on the other hand, heatmaps create a disconnection from the overall estimation of the keypoints, making the intermediate clustering and grouping process not directly integrable to be end-to-end learning frameworks. 

\textbf{Regression-based} methods \cite{Carreira2016HumanPE,zhou2019objects,nie2019single,wei2020point,sun2018integral} aim to directly approach keypoint detection with a direct loss minimization between predicted and ground truth coordinates, hence, they can be more easily integrated into an end-to-end learning framework. However, holistic regression can be intrinsically more difficult to optimize due to the high-precision needed by pose recognition. Furthermore, regression-based approaches typically have a recursive procedure \cite{dollar2010cascaded} that skips a large number of candidate locations, creating a performance gap with the heatmap-based methods. Our work follows the line of regressive pose estimation, and formulates the process of step-by-step regression \cite{dollar2010cascaded,Carreira2016HumanPE} implicitly in a layered Transformer way. 

\textbf{Transformers and self-attention} 
The attention mechanism \cite{xu2015show,vaswani2017attention,devlin2018bert} has greatly advanced the field of representation learning in machine learning. The introduction of Transformers \cite{vaswani2017attention} to  object detection gives another leap-forward in building end-to-end object detection framework that is free of proposal, anchor, and post processing (non-maximum suppression). Here, we build cascade Transformers based on the DETR \cite{carion2020end} framework to perform regression-based pose recognition. Our system, named PRTR, aims towards a general-purpose keypoint regression solution without specific heuristic-driven designs.

Recently, Transformer architecture and self-attention have seen increasing application in computer vision tasks \cite{Parmar2018ImageT, carion2020end, Dosovitskiy2020AnII}, yet there are limited visualization works compared with those done on language application \cite{Coenen2019VisualizingAM, Vig2019AnalyzingTS}. As far as we know, we are the first to visualize the dynamic decoding process in Transformer decoder, which brings significant insights to future Transformer designs.

\section{Method}

\begin{figure*}[!t]
\centering
\includegraphics[width=0.85\linewidth]{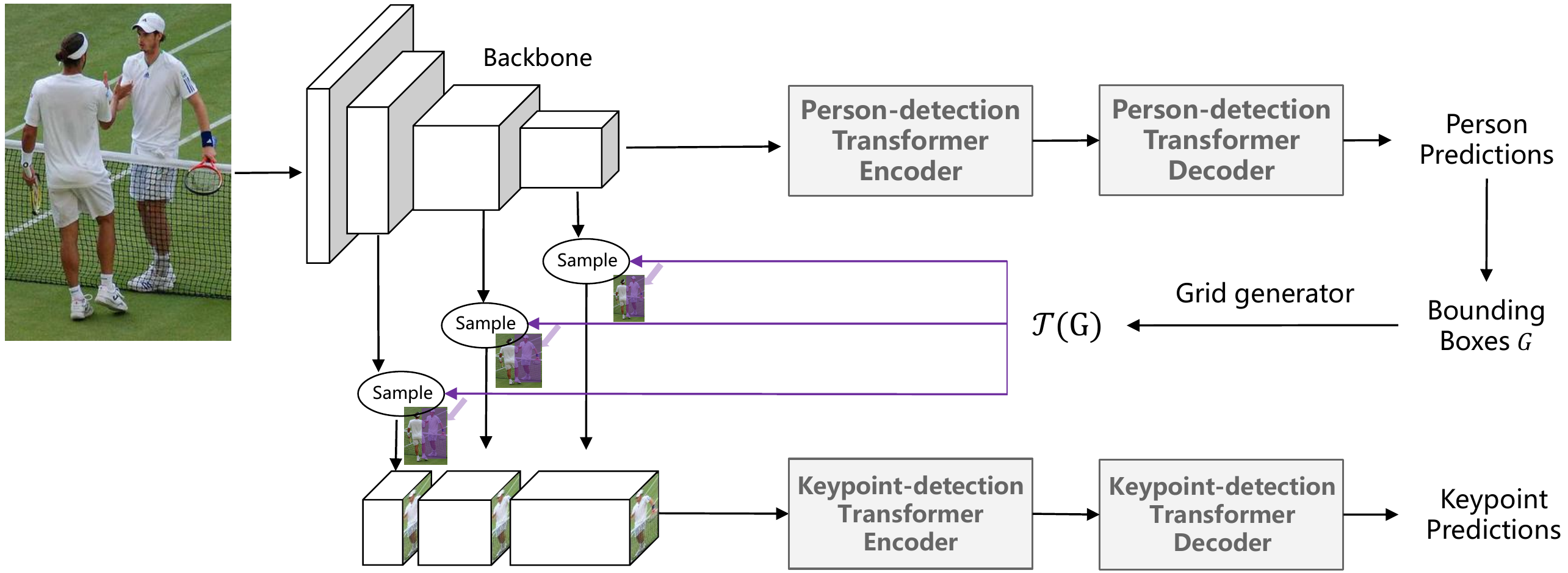}
\vspace{-1mm}
\caption{The architecture of Pose Recognition with TRansformer (\textbf{PRTR}), \textbf{end-to-end variant}. For end-to-end learning, instead of cropping at RGB image level, we apply differentiable bilinear sampling on multiple layers of backbone-generated features to provide \emph{zoomed-in} and \emph{multi-level} feature for keypoint-detection Transformer.}
\label{fig:model2}
\vspace{-3mm}
\end{figure*}

We argue that the attention mechanism in Transformer can act as a general-purpose inference engine for regression in vision tasks by writing visual perception as a Bayesian inference $P(Y|I) \propto P(I|Y) P(Y)$ with  $Y=(\hat{y}_{elb},\,\hat{y}_{knee},\cdots,\,\hat{y}_{nose})$. Here, Transformer for regression performs direct learning and inference by capturing complex joint relations between input $I$ and prediction hypotheses (queries), $P(I|Y)$,  through cross-attention, and modeling the prior on configuration of $Y$, $P(Y)$,  via hypothesis (query) self-attention. See Figure~\ref{teaser}.

In this section, we instantiate this idea as Pose Recognition with TRansformer (PRTR) for multi-person pose recognition. The overall architecture is shown in Figure~\ref{fig:model1}. We first introduce a cascaded double Transformer architecture for person and keypoint detection, then an end-to-end variant to streamline the entire model.

\vspace{-2mm}
\subsection{Person-Detection Transformer}
\vspace{-2mm}
We tackle multi-person pose recognition problem in a top-down manner, and adopt a Transformer architecture \cite{vaswani2017attention} following DEtection TRansformer (DETR) \cite{carion2020end} as the backbone for the first-stage person detection. In the encoder stage, image features generated by a CNN are flattened and fed into a Transformer encoder to produce contextualized image features; in the decoder stage, given a fixed set of learned query embedding as input, Transformer decoder reasons about the relations between objects under the context of image features, and output all the object queries in a parallel way. At last, a classification head is used to classify the object as person or background ($\varnothing$), and a 4-channel regression head is used to predict the bounding boxes.

\subsection{Keypoint-Detection Transformer}

After getting the bounding boxes, we crop the RGB image and use another CNN backbone to get feature maps per person. Because only matched queries are involved in calculating the loss for keypoint-detection Transformer, we filtered out unmatched ones. Like the process of person detection, we use the encoder-decoder architecture of the Transformer to predict in a parallel fashion, but we use another set of queries (quantity denoted $Q$). Finally, a classification head predicts among $J$ types of joints and background ($\varnothing$) and a 2-channel regression head outputs the coordinate of each keypoint. 

Since PRTR infers a fixed larger number of predictions than ground truth (quantity denoted $J$), we need to find a matching between them to calculate the loss. We formulate this matching problem as an \emph{optimal bipartite matching} problem, which can be solved efficiently by Hungarian algorithm \cite{Stewart2016EndtoEndPD}. In specific, we try to find an injective function $\sigma \in [J] \rightarrow [Q] $ that firstly minimizes the matching cost $\mathcal{C}$ in a discrete way:

\vspace{-4mm}
\begin{align}
\mathcal{C}=\underset{\sigma}{\arg \min } \sum_{i}^{J} \mathcal{C}\left(y_{i}, \hat{y}_{\sigma(i)}\right)
\end{align}
~, where $\hat{y}_{\sigma(i)}$ means the prediction to be matched with the $i$-th keypoint.

At training stage, we match our queries using a mixture of classification probabilities and coordinate deviation. For instance, the cost function for the $i$-th keypoint and its matched query $\sigma(i)$ is:
\vspace{-2mm}
\begin{align}
\mathcal{C}_{i} = - \hat{p}_{\sigma(i)}(c_i) + 
\| b_i - \hat{b}_{\sigma(i)} \| \label{eqn:hungarian_cost_train}
\end{align}
~, where $\hat{p}_{\sigma(i)}$ is the class probabilities of the query and $c_i$ is the class label for $i$-th keypoint. However, at inference stage, we do not have access to the ground-truth keypoint coordinates, thus we match $J$ prototype keypoints to queries using only the classification probabilities. Therefore the matching cost for $i$-th keypoint is simply:
\vspace{-1mm}
\begin{align}
\mathcal{C}_{i} = - \hat{p}_{\sigma(i)}(c_i) \label{eqn:hungarian_cost_test}
\end{align}

After running the bipartite matching algorithm, we return the matched $J$ keypoints as our prediction.

The loss function of the model is obtained by replacing negative probabilities in Equation~\ref{eqn:hungarian_cost_train} with negative log-likelihood $-\log \hat{p}_{\sigma(i)}(c_i)$ for matched queries. For unmatched queries we only backpropagate the classification loss. To address the class imbalance caused by $\varnothing$ class, as in \cite{carion2020end}, we set the weight of its log-probability term to 0.1.

\subsection{Multi-layer Cropping with STN}

\begin{table*}[!th]
\centering
\caption{Comparisons on COCO \textbf{val} set. $^+$ indicates using multi-scale test. $^\ast$ indicates the end-to-end model variant. 
}
\label{tab:coco_val}
\vspace{-4pt}
\begin{adjustbox}{max width=.95\linewidth}
\begin{tabular}{l|l|c|c|c|cccccccc} \hline
Method  & Backbone     & Input size & \#Params & GFLOPs & $AP$      & $AP_{50}$ & $AP_{75}$ & $AP_M$  & $AP_L$  & $AR$   \\ \hline
\multicolumn{11}{c}{Heatmap based} \\ \hline
8-stage Hourglass \cite{newell2016stacked} & Hourglass-8 stacked & 256 $\times$ 192 & 25.1M & 14.3 & 66.9 & $-$ & $-$ & $-$ & $-$ & $-$ \\
CPN \cite{chen2018cascaded} & ResNet-50 & 256 $\times$ 192 & 27.0M & 6.20 & 68.6 & $-$ & $-$ & $-$ & $-$ & $-$ \\
SimpleBaseline \cite{xiao2018simple} & ResNet-50 & 384 $\times$ 288 & 34.0M & 18.6 & 72.2 & 89.3 & 78.9 & 68.1 & 79.7 & 77.6 \\
SimpleBaseline \cite{xiao2018simple} & ResNet-101 & 384 $\times$ 288 & 53.0M & 26.7 & 73.6 & 89.6 & 80.3 & 69.9 & 81.1 & 79.1 \\
HRNet \cite{sun2019deep} & HRNet-W32 & 384 $\times$ 288 & 28.5M & 16.0 & 75.8 & 90.6 & 82.7 & 71.9 & 82.8 & 81.0 \\ 
\hline
\multicolumn{11}{c}{Regression based} \\ \hline
PointSetNet$^+$ \cite{wei2020point} & ResNeXt-101-DCN &  $-$ & $-$ & $-$ & 65.7 & 85.4 & 71.8 & $-$ & $-$ & $-$ \\
PointSetNet$^+$ \cite{wei2020point} & HRNet-W48 &  $-$ & $-$ & $-$ & 69.8 & 88.8 & 76.3 & $-$ & $-$ & $-$ \\
\hline
PRTR$^\ast$ & ResNet-101 & $-$ & $-$ & $-$ & 64.8 & 85.1 & 70.2 & 60.4 & 73.8 & 73.9 \\
PRTR$^\ast$ & HRNet-W48  & $-$ & $-$ & $-$ & 66.2 & 85.9 & 72.1 & 61.3 & 74.4 & 72.2 \\
PRTR & ResNet-50 & 384 $\times$ 288 & 41.5M & 11.0 & 68.2 & 88.2 & 75.2 & 63.2 & 76.2 & 76.0 \\
PRTR & ResNet-50 & 512 $\times$ 384 & 41.5M & 18.8 & 71.0 & 89.3 & 78.0 & 66.4 & 78.8 & 78.0 \\
PRTR & ResNet-101 & 384 $\times$ 288 & 60.4M & 19.1 & 70.1 & 88.8 & 77.6 & 65.7 & 77.4 & 77.5 \\
PRTR & ResNet-101 & 512 $\times$ 384 & 60.4M & 33.4 & 72.0 & 89.3 & 79.4 & 67.3 & 79.7 & 79.2 \\
PRTR & HRNet-W32 & 384 $\times$ 288 & 57.2M & 21.6 & 73.1 & {\bf 89.4} & 79.8 & 68.8 & 80.4 & 79.8 \\
PRTR & HRNet-W32 & 512 $\times$ 384 & 57.2M & 37.8 & {\bf 73.3} & 89.2 & {\bf 79.9} & {\bf 69.0} & {\bf 80.9} & {\bf 80.2} \\
\hline
\end{tabular} 
\end{adjustbox}
\vspace{-3mm}
\end{table*}

In the previous section, we introduce a two-stage pipeline. However, under an end-to-end philosophy, it is desired that the model is end-to-end tunable to exploit the synergy between person detection and keypoint recognition task. To this end, we incorporate the Spatial Transformer Network (STN) \cite{Fang2019LocalityConstrainedST} to crop out image features needed by the keypoint-detection Transformer directly from the feature map generated by the first CNN backbone. This cropping operation is differentiable not only to the feature maps, but also to the bounding box coordinates. 

For instance, an $w \times h$ grid generated by $b = (x_{\text{left}}, x_{\text{right}}, y_{\text{top}}, x_{\text{down}})$ can be formulated by: 
\begin{align}
x_i &= \frac{w-i}{w} x_{\text{left}} + \frac{i}{w} x_{\text{right}} \\
y_j &= \frac{h-j}{h} y_{\text{top}} + \frac{j}{h} y_{\text{down}}
\end{align}
, where $b$ is relative to the original image, and $w \times h$ is the desired feature map size for the keypoint-detection Transformer.

To mitigate the resolution challenge commonly seen in keypoint recognition, we apply the grid to feature maps of different scales generated at different intermediate layers of the CNN backbone using a bilinear kernel. Denoting the the original $W \times H$ feature map by $U$, the differentiable sampling process can be formulated as:
\begin{equation}
\resizebox{0.9\hsize}{!}{%
    $\displaystyle V_{ij}=\sum_{m,n}
    U_{nm} \max \left(0,1-\left|x_{i}-m\right|\right) \max \left(0,1-\left|y_{j}-n\right|\right)$
    }
\end{equation}

After getting a series of image features of the same spatial size, we concatenate them into a single feature map for the keypoint-detection Transformer. This multi-layer cropping variant is illustrated in Figure~\ref{fig:model2}. 

\section{Experiment}

We validate our proposed method on the COCO Keypoint Detection task and MPII Human Pose Dataset. 

\subsection{Experiment Setup}

\begin{table*}[!th]
\centering
\caption{Comparisons on COCO \textbf{test-dev} set, excluding systems trained with external data. $^+$ means using multi-scale test. $^\ast$ means end-to-end model variant. For bottom-up methods and end-to-end PRTR, computation overheads are not shown for being incomparable to two-stage methods. \#Params and FLOPs are calculated for the pose estimation network, excluding human detection and keypoint grouping. Table format is adapted from \cite{wei2020point} and \cite{sun2019deep}.}
\label{t1}
\vspace{-4pt}
\begin{adjustbox}{max width=.95\linewidth}
\begin{tabular}{l|l|c|c|c|cccccccc} \hline
Method  & Backbone     & Input size & \#Params & GFLOPs & $AP$      & $AP_{50}$ & $AP_{75}$ & $AP_M$  & $AP_L$  & $AR$   \\ \hline
\multicolumn{11}{c}{Heatmap based: keypoint heatmap prediction and post-processing to decode coordinates} \\ 
\hline
CMU-Pose \cite{cao2017realtime}  & 3CM-3PAF  & $-$ & $-$ & $-$  & 61.8    & 84.9      & 67.5      & 57.1    & 68.2    & 66.5 \\  
Mask-RCNN \cite{he2017mask} & ResNet-50 &  $-$ & $-$ & $-$ &  63.1 & 87.3 & 68.7 & 57.8 & 71.4 & $-$ \\
G-RMI \cite{papandreou2017towards} & ResNet-101 & 353 $\times$ 257 &  42.6M & 57.0 & 64.9 & 85.5 & 71.3 & 62.3 & 70.0 & 69.7 \\
Assoc. Embed.  \cite{newell2017associative} & Hourglass-4 stacked  & $-$ & $-$ & $-$ & 65.5    & 86.8      & 72.3      & 60.6    & 72.6    & 70.2 \\
PifPaf  \cite{kreiss2019pifpaf}   & ResNet-101-dilation  &  $-$ & $-$ & $-$ & 66.7    & $-$ & $-$      & 62.4    & 72.9    & $-$    \\
PersonLab \cite{papandreou2018personlab}   & ResNet-101  &  $-$ & $-$ & $-$ & 65.5    & 87.1      & 71.4      & 61.3    & 71.5    & 70.1 \\
PersonLab$^+$        & ResNet-101 &  $-$ & $-$ & $-$  & 67.8    & 88.6      & 74.4      & 63.0    & 74.8    & 74.5 \\
HigherHRNet$^+$ \cite{cheng2020higherhrnet} & HRNet-W48 &  $-$ & $-$ & $-$ & 70.5 & 89.3 & 77.2 & 66.6 & 75.8 & 74.9 \\
CPN \cite{chen2018cascaded} & ResNet-Inception & 384 $\times$ 288 & $-$ & $-$ & 72.1 & 91.4 & 80.0 & 68.7 & 77.2 & 78.5 \\
SimpleBaseline \cite{xiao2018simple} & ResNet-152 & 384 $\times$ 288 & 68.6M & 35.6 & 73.7 & 91.9 & 81.1 & 70.3 & 80.0 & 79.0 \\
HRNet \cite{sun2019deep} & HRNet-W48 & 384 $\times$ 288 & 63.6M & 32.9 & 75.5 & {\bf 92.5} & 83.3 & 71.9 & 81.5 & 80.5 \\
DARK \cite{zhang2020distribution} & HRNet-W48 & 384 $\times$ 288 & 63.6M & 32.9 & {\bf 76.2} & {\bf 92.5} & {\bf 83.6} & {\bf 72.5} & {\bf 82.4} & {\bf 81.1} \\
\hline
\multicolumn{11}{c}{Regression based: direct keypoint coordinate prediction} \\
\hline
CenterNet$^+$  \cite{zhou2019objects} & Hourglass-2 stacked  &  $-$ & $-$ & $-$ & 63.0    & 86.8       & 69.6     & 58.9    & 70.4    & $-$    \\  
DirectPose \cite{Tian2019DirectPoseDE}   & ResNet-101  &  $-$ & $-$ & $-$ & 63.3    & 86.7       & 69.4     & 57.8    & 71.2    & $-$    \\  
SPM$^+$ \cite{nie2019single}   & Hourglass-8 stacked & 384 $\times$ 384 & $-$ & $-$ & 66.9    & 88.5       & 72.9     & 62.6    & 73.1    & $-$    \\  
Integral \cite{sun2018integral}  & ResNet-101 & 256 $\times$ 256 & 45.0M & 11.0  & 67.8    & 88.2       & 74.8     & 63.9    & 74.0    & $-$    \\
PointSetNet$^+$ \cite{wei2020point} & HRNet-W48 &  $-$ & $-$ & $-$  & 68.7    & 89.9   & 76.3     & 64.8    & 75.3    & $-$    \\  
\hline
PRTR$^\ast$     & ResNet-101 &  $-$ & $-$ & $-$  & 63.4    & 86.2       & 69.4     & 59.3    & 72.0    & 73.0 \\  
PRTR$^\ast$    & HRNet-W48 &  $-$ & $-$ & $-$ & 64.9    & 87.0       & 71.7     & 60.2    & 72.5    & 74.1 \\
PRTR & ResNet-101 & 384 $\times$ 288 & 60.4M & 19.1 & 68.8   &  89.9  & 76.9 & 64.7  &  75.8  &  76.6 \\ 
PRTR & ResNet-101 & 512 $\times$ 384 & 60.4M & 33.4 & 70.6 & 90.3 & 78.5 & 66.2 & 77.7 & 78.1 \\
PRTR & HRNet-W32 & 384 $\times$ 288 & 57.2M & 21.6 & 71.7 & {\bf 90.6} & {\bf 79.6} & 67.6 & 78.4 & 78.8 \\
PRTR & HRNet-W32 & 512 $\times$ 384 & 57.2M & 37.8 & {\bf 72.1} & 90.4 & {\bf 79.6} & {\bf 68.1} & {\bf 79.0} & {\bf 79.4} \\ \hline \end{tabular}
\end{adjustbox}
\end{table*}

\begin{table*}[!ht]
\centering
\vspace{2mm}
\caption{Comparisons on the MPII \textbf{val} set (PCKh@0.5). }
\label{t2}
\vspace{-2mm}
\begin{adjustbox}{max width=0.8\linewidth}
\begin{tabular}{l|l|ccccccc|c} \hline
Method               & Backbone   & Head & Sho & Elb & Wri & Hip & Knee & Ank & Mean \\ \hline
\textbf{\emph{Heatmap Based}}  &&&&&&&& \\
Convolutional Pose Machines \cite{wei2016convolutional} & CPM & 96.2 & 95.0 & 87.5 & 82.2 & 87.6 & 82.7 & 78.4 & 87.7 \\
Simple Baseline \cite{xiao2018simple} & ResNet-152 & 97.0 & 95.9 & 90.3 & 85.0 & 89.2 & 85.3 & 81.3 & 89.6 \\
HRNet \cite{sun2019deep} & HRNet-W32 & {\bf 97.1} & {\bf 95.9} & {\bf 90.3} & {\bf 86.4} & {\bf 89.1} & {\bf 87.1} & {\bf 83.3} & {\bf 90.3} \\ 
\hline
\textbf{\emph{Regression Based}}  &&&&&&&& \\
Integral \cite{sun2018integral} & ResNet-101 & $-$ & $-$ & $-$ & $-$ & $-$ & $-$ & $-$ & 87.3 \\
PRTR (ours)  & ResNet-101 & 96.3 & 95.0 & 88.3 & 82.4 & 88.1 & 83.6 & 77.4 & 87.9   \\ 
PRTR (ours)  & ResNet-152 & 96.4 & 94.9 & 88.4 & 82.6 & 88.6 & 84.1 & 78.4 & 88.2  \\
PRTR (ours) & HRNet-W32 & {\bf 97.3} & {\bf 96.0} & {\bf 90.6} & {\bf 84.5} & {\bf 89.7} & {\bf 85.5} & {\bf 79.0} & {\bf 89.5} \\ \hline
\end{tabular}
\end{adjustbox}
\end{table*}
\vspace{-1mm}

\textbf{Datasets.} We used two human pose estimation datasets, COCO and MPII. The COCO dataset \cite{lin2014microsoft} contains over 200,000 images and 250,000 person instances. Each person instance is labelled with 17 joints. We train our model on COCO train2017 dataset with 57K images, and evaluate our approach on the standard val2017 and test-dev2017 split, containing 5K and 20K images respectively. The MPII single person dataset \cite{andriluka20142d} consists of around 25K images and 40K well-separated person instances. We follow the standard train/val split.

\textbf{Evaluation metrics.} We follow the common practice in \cite{sun2019deep} and use Object Keypoint Similarity (OKS) for COCO and Percentage of Correct Keypoints (PCK) for MPII to evaluate the performance.

\textbf{Person-detection Transformer finetuning.} We first tune a person detector by initializing from weights provided by DETR \cite{carion2020end}. We keep all weights except prototype vectors for non-person class in the classifier. The tuning lasts for 10 epochs with a leaning rate of 1e$-$7 for ResNet-50 backbone and 5e$-$6 for the rest. For pose recognition task, people without any visible keypoints are not desired to be detected; these people have a common characteristic of being small in area. In fact, all people with a segmentation area less than $32^2$ do not contain keypoints. Given this, we skipped person annotations without visible keypoints at this stage for both training and evaluation. After tuning, the person detector scores an mAP of $67.0$ on the pruned val2017 set, and an mAP of $50.2$ on the standard val2017 set.  

\textbf{Two-stage variant.} For the two-stage version of our model, we extend the human detection bounding box in height or width to a fixed aspect ratio ($4:3$ for COCO). A patch is cropped using the box and then resized to a fixed size, $384 \times 288$ or $512 \times 384$ for COCO. The data augmentation follows \cite{xiao2018simple}, including random rotation ($[-40^\circ, 40^\circ]$), random scale ($[0.7, 1.3]$), and flipping. The data pre-processing remains the same for MPII, except for aspect ratio set to $1:1$ and input size available in $384 \times 384$ or $512 \times 512$. For the Transformer part, number of encoder layers, decoder layers and keypoint queries are set to 6, 6, 100 respectively.

We use the AdamW optimizer \cite{Loshchilov2019DecoupledWD}. The base learning rate is 1e$-$5 for ResNet backbone and 1e$-$4 for the rest, with weight decay 1e$-$4. Multi-step learning rate schedule is used, which halves the learning rate at the 120th and 140th epoch respectively. The training process terminates within 200 epochs for both datasets.

\textbf{Testing.}
At test time, We use the person detection results from the tuned person detector (with AP $50.2$ on COCO val2017 set) for both COCO val and test-dev set. Inspired by the common practice of flip-test \cite{chen2018cascaded, newell2016stacked, xiao2018simple} used in heatmap paradigms, we compute the keypoint coordinates by averaging the outputs of original and flipped images.

\textbf{End-to-end variant.} For the end-to-end variant, we use ground truth to match predicted people after person-detection Transformer, and discard unmatched queries because they will not be contributing to training keypoint-detection Transformer. For images with more than 5 people, we randomly sample 5 matched queries to reduce computational cost. Bounding boxes predicted by person-detection Transformer are enlarged by $25\%$ at both the height and width dimension before sampling image features from backbone features, which helps predicting keypoints at the margin by taking in more contextual information. 

We used the same data augmentation as DETR \cite{carion2020end} except randomly resizing the image to having its shortest side being $760$ to $1024$ while not exceeding $1400$. Optimizer settings follow the two-stage variant, except for halving the learning rate at the 25th and 60th epoch instead.

\vspace{-1mm}
\subsection{Results}
\vspace{-1mm}

\begin{figure*}[!ht]
\centering
\includegraphics[width=\linewidth]{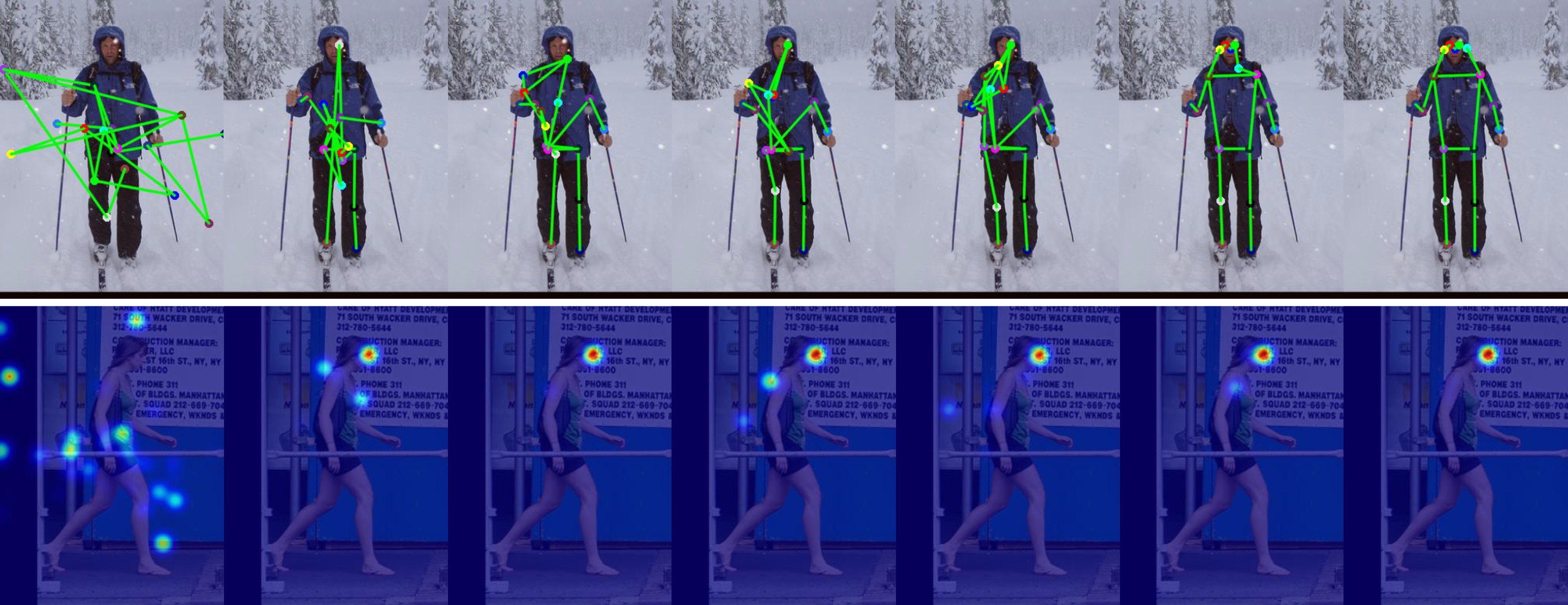}
\vspace{-2mm}
\caption{\small Visualization of PRTR's decoding process for the keypoint detection Transformer. 
In the first row, the last column shows the final predictions and the former 6 columns show the predictions for the initial query embedding and the intermediate 5 decoder layers. The second row shows an overlay of heatmaps of 100 queries for Right Ear and Left Eye respectively.
}
\label{ffig5}
\vspace{-3mm}
\end{figure*}

\begin{figure}[!t]
    \centering
    \begin{subfigure}{\linewidth}
    \centering
    \includegraphics[height=4.1cm]{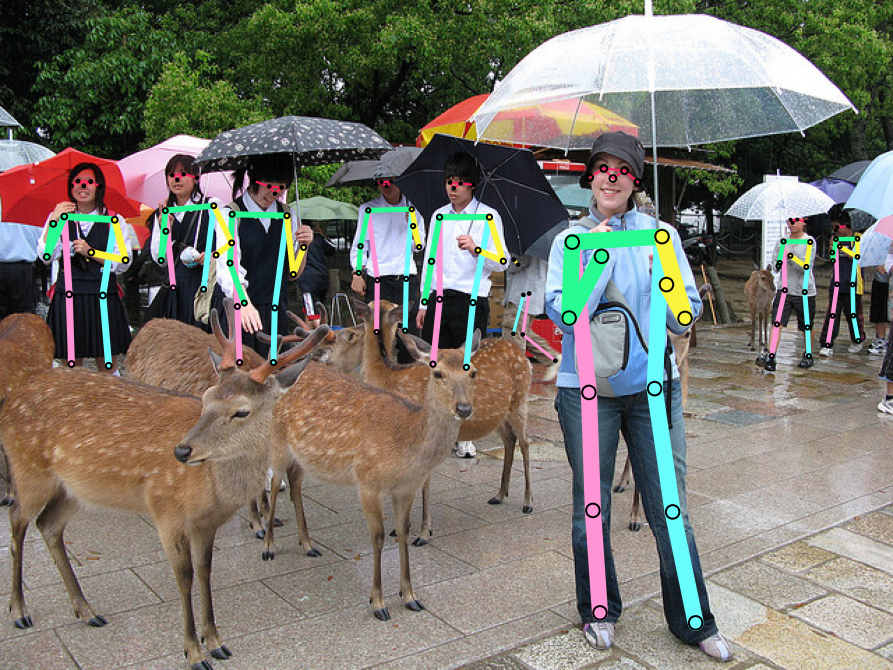}
    \includegraphics[height=4.1cm]{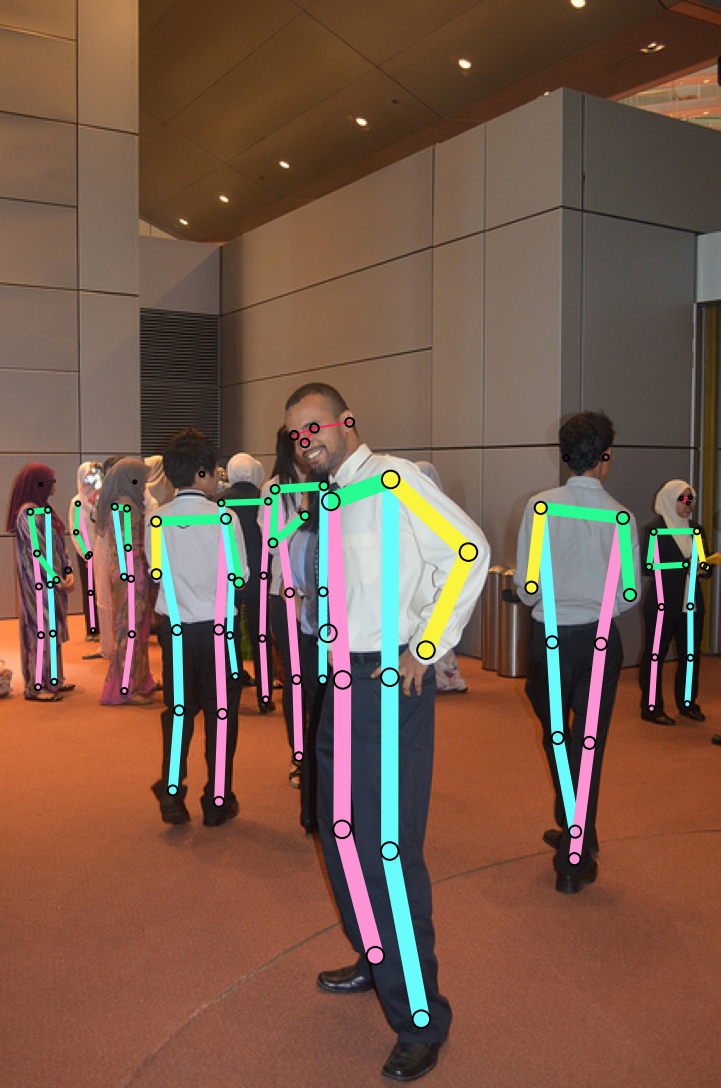}
    \end{subfigure}\\
    \begin{subfigure}{\linewidth}
    \centering
    \includegraphics[height=3.075cm,trim=2cm 0 0 0,clip]{poses/1.png}
    \includegraphics[height=3.075cm]{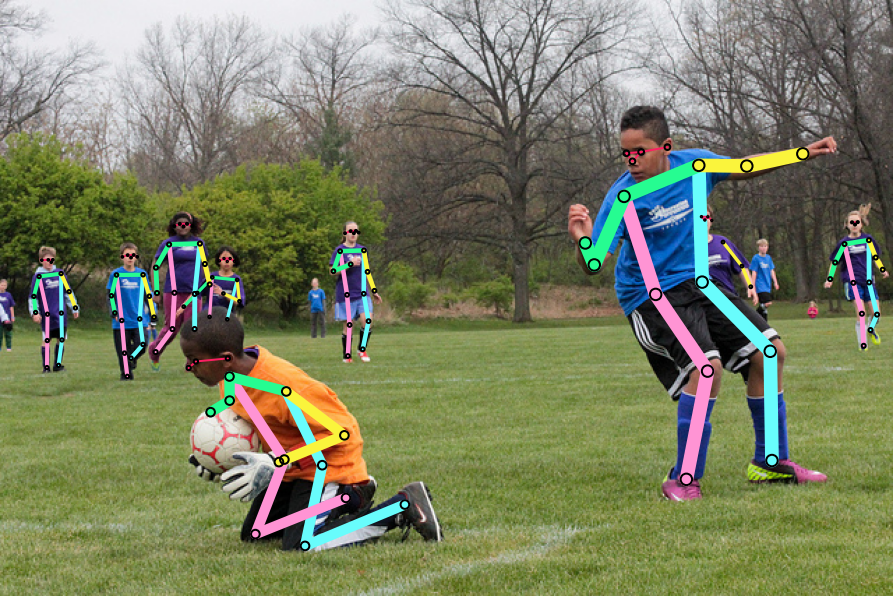}
    \end{subfigure}\\
    \caption{Qualitative COCO human pose estimation results on images of varying sizes and poses.}
    \label{fig:pose_vis}
\vspace{-3mm}
\end{figure}

\textbf{Results on the COCO dataset.} Table~\ref{tab:coco_val} and Table~\ref{t1} compare pose estimation results on COCO val and test-dev set respectively. Qualitative results are given in Figure~\ref{fig:pose_vis}. For the end-to-end variant, it surpasses competing fully end-to-end components like CenterNet \cite{zhou2019objects} and DirectPose \cite{Tian2019DirectPoseDE}. The two-stage variant of our approach outperforms the competing baselines in the \textit{regression based} category. Our model with ResNet-101 backbone is comparable to PointSetNet \cite{wei2020point} which leverages a more complex backbone (HRNet-W48). Our model benefits from larger input size and stronger feature backbones. By enlarging input size from $384 \times 288$ to $512 \times 384$, PRTR with ResNet-50 and ResNet-101 receives 2.2, 1.9 improvement respectively. Our best model, achieving 72.1 AP, is able to emulate the heatmap-based HigherHRNet \cite{cheng2020higherhrnet}.

\textbf{Results on the MPII val dataset.} Since only MPII val is publicly available, we report the performance of our model trained on the entire MPII train set, as shown in Table~\ref{t2}. Our best model achieves a 89.5 PCKh@0.5 score, comparable to that of SimpleBaseline \cite{xiao2018simple}. Not needing a person detection stage, MPII is not tried with the end-to-end variant.

\subsection{Ablation Studies}

We perform ablation studies on COCO dataset to verify our design choices as listed in Table~\ref{tbl_abl_1} and \ref{tbl_abl_2}. The results presented are on COCO val2017, with ResNet-50 backbone and input size ~$384 \times 288$.

\begin{table}[!t]
\centering
\caption{Ablation study \wrt number of queries on COCO val2017. \emph{Fixed} stands for class-specific queries, \ie, a query is always mapped to a fixed keypoint type.}
\label{tbl_abl_1}
\vspace{-4pt}
\scalebox{0.93}{
\begin{tabular}{l|cccccc} 
\hline
\#Queries  & $AP$      & $AP_{50}$ & $AP_{75}$ & $AP_M$  & $AP_L$  & $AR$   \\
\hline
100 & {\bf 67.7} & 87.7 & {\bf 74.9} & 62.6 & {\bf 75.7} & {\bf 74.2} \\
50 & 67.6 & 87.7 & 74.8 & {\bf 63.0} & 75.4 & 74.1 \\
17 & 67.3 & {\bf 87.9} & 74.4 & 62.1 & 75.4 & 73.1 \\
17 (Fixed) & 56.3 & 83.7 & 61.9 & 54.2 & 60.3 & 69.6 \\\hline
\end{tabular}
}
\vspace{-6mm}
\end{table}

\textbf{Non class-specific queries.} We make the queries of Transformer decoder to predict both keypoint coordinates and classes, and then select the required points from all the queries via class probabilities. This way, we do not enforce a fixed correspondence between $J$ keypoint types and queries. Therefore, the queries are not class-specific and can be used to predict different types of keypoints each time. Here, we focus on two alternative designs: a) different number of queries used; b) when number of queries equals the number of required points, the necessity for queries to be non class-specific. From Table~\ref{tbl_abl_1}, it is clear that 100-query version only has a small advantage over 50- and 17-query counterparts. However, using class-specific queries will greatly hamper the performance of the model, resulting in a large drop in AP (11.4). This illustrates the necessity that each query dynamically predicts its preferred keypoint type, and reads out the best estimation through Hungarian matching during inference.

\begin{table}[!t]
\centering
\caption{Ablation study on COCO val2017. 'GT Box', '$\varnothing$ Logit' represent ground truth box for cropping, and inclusion of background logits during inference respectively.}
\label{tbl_abl_2}
\vspace{-4pt}
\setlength{\tabcolsep}{2pt}
\begin{tabular}{ccc|cccccc} 
\hline
\bigcell{c}{GT \\ Box} & \bigcell{c}{$\varnothing$ \\ Logit}  & \bigcell{c}{Flip \\ Test} & $AP$      & $AP_{50}$ & $AP_{75}$ & $AP_M$  & $AP_L$  & $AR$   \\
\hline
& &  & 67.1 & 87.6 & 74.5 & 62.6 & 74.7 & 73.7 \\
\cmark & & & 69.1 & 90.1 & 77.0 & 66.1 & 73.7 & 73.9 \\
& \cmark & & 66.2 & 87.2 & 73.5 & 62.1 & 72.8 & 72.8 \\
\cmark & \cmark & & 68.2 & 89.7 & 75.5 & 65.3 & 72.5 & 72.9 \\
& & \cmark & 67.7 & 87.7 & 74.9 & 62.6 & 75.7 & 74.2 \\
\cmark & & \cmark & 70.4 & 91.2 & 78.3 & 67.1 & 75.2 & 74.7 \\
& \cmark & \cmark & 66.4 & 86.9 & 73.0 & 62.0 & 73.4 & 72.8 \\
\cmark & \cmark & \cmark & 68.9 & 89.9 & 75.8 & 65.7 & 73.4 & 73.2 \\ \hline
\end{tabular}
\vspace{-4mm}
\end{table}

\textbf{Exclusion of background prediction during inference.} During inference, we exclude the logits of the background class ($\varnothing$) before normalizing class probabilities to provide more keypoint candidates for the Hungarian matcher. From Table~\ref{tbl_abl_2}, we observe that including the logits of background class will result in a 0.9$-$1.5 drop in AP.

\begin{figure*}[!t]
\centering
\includegraphics[width=\linewidth]{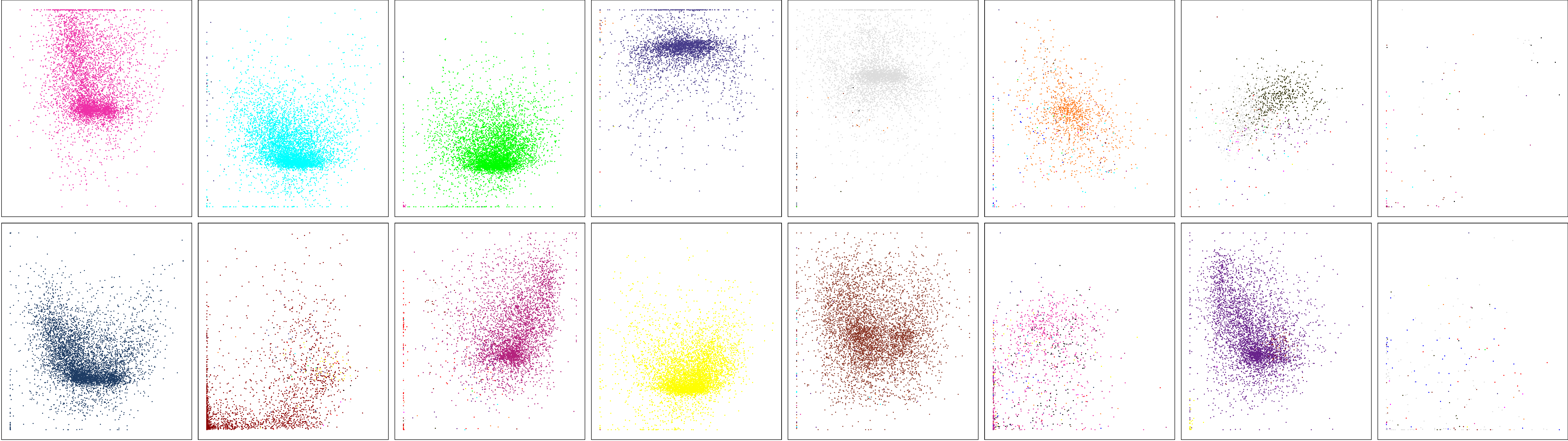}
\caption{\small Visualization of 16 keypoint (excluding the background class) prediction out of $Q=100$ queries in the keypoint-detection Transformer on COCO val2017. Each colored dot represents a predicted keypoint for the corresponding class.
}
\label{query_heatmap}
\vspace{-2mm}
\end{figure*}

\textbf{Flip test.} Flipping is a common test augmentation used in heatmap paradigms, where input image is horizontally flipped and fed to the model, and then flip back, align and average the predicted heatmaps to increase accuracy. The same technique applies to regression models as well, with results obtained by directly averaging the predicted keypoint coordinates. Since regression operates on continuous coordinate space, one advantage is that it does not suffer from the inaccuracy caused by alignment errors in heatmap paradigms, as described in \cite{Huang_2020_CVPR}. From Table~\ref{tbl_abl_2}, flip test offers a consistent performance boost for our model.

\textbf{Oracle results.} We also explore the room for improvement by replacing the bounding boxes predicted by person-detector with ground truth (GT) ones, as in Table~\ref{tbl_abl_2}. It is evident that GT boxes improves AP by 2$-$2.5, indicating the potential benefit of a stronger person-detector.

\subsection{Vis. for Keypoint Detection Transformer}

\begin{figure}[!ht]
\centering
\includegraphics[width=7.5cm]{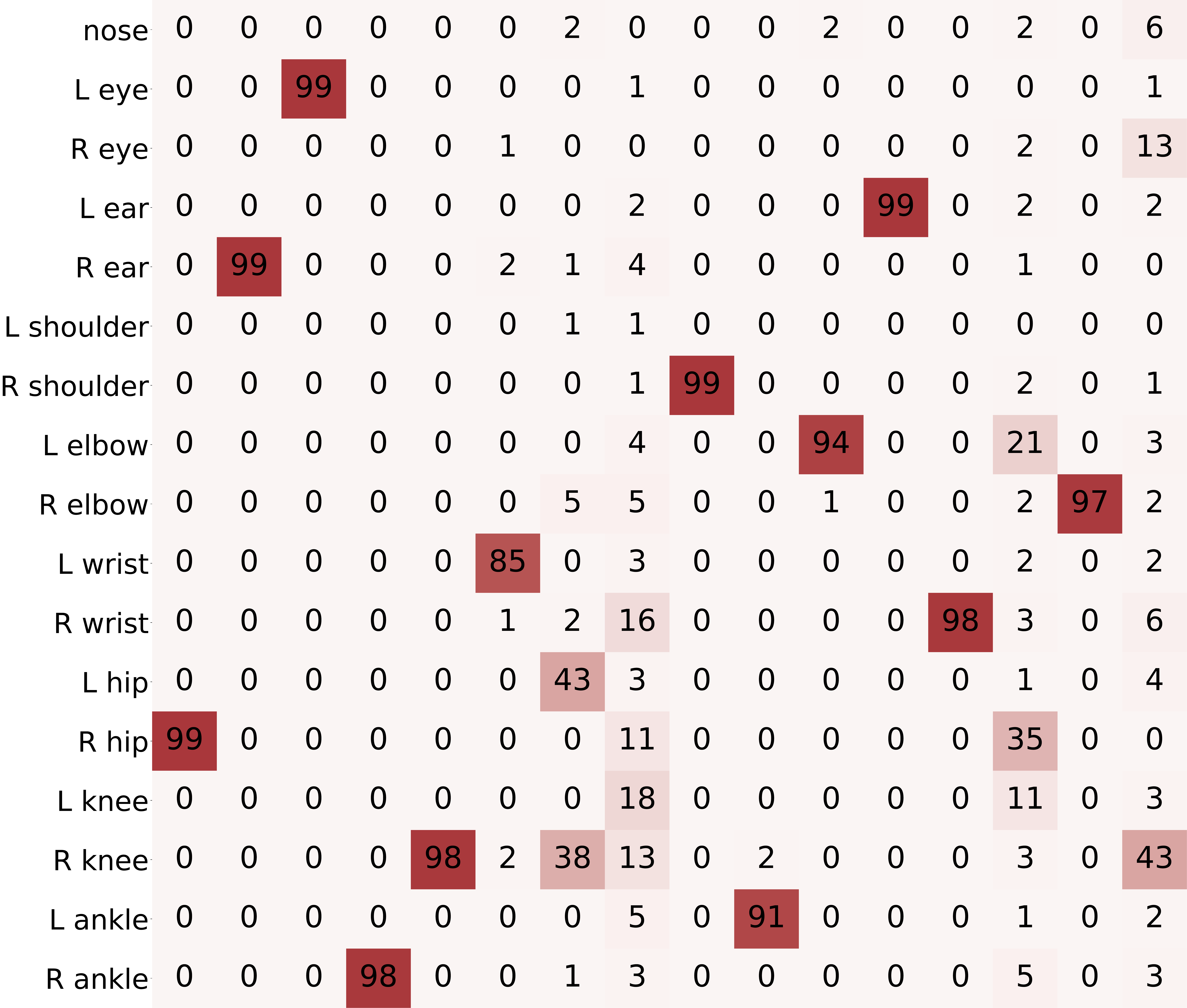}
\caption{Visualization of distributions of predicted keypoint classes for 16 out of a total of $Q=100$ queries in the keypoint-detection Transformer. Numbers on heatmap correspond to the probability ($\times 100$) for the individual keypoint classes. We observe that queries learn to specialize on keypoint classes.}
\label{query_dist}
\vspace{-3mm}
\end{figure}

In this section, we show visualizations for the keypoint detection Transformer. 
In Figure~\ref{query_heatmap} and Figure~\ref{query_dist} we visualize the position and class distribution for keypoint predictions by the queries. Different queries are observed to bias towards different keypoints (\eg in our model 92.3\% of the predictions by the 89th query are nose keypoints). We also observe that queries dedicated to certain keypoints are biased to specific locations (\eg the query focusing on the nose tends to predict positions in the upper part of the images) while the points predicted by queries focusing on background are uniformly distributed.

In Figure~\ref{ffig5}, we explore and visualize query output results in different decoder layers during inference. The first row shows the queries selected by the Hungarian algorithm and demonstrate how their predictions move and refine through lower-to-higher decoder layers. Initially, the predictions are randomly located in the image. After passing some decoder layers, queries predictions gradually approach the proper locations. It is noteworthy that if a query's prediction is close to the ground truth in lower layers, its prediction barely changes in higher layers.

The second row shows the spatial probabilities of a certain type of keypoint. For visualization, Gaussian heatmaps are first generated around the predicted keypoint locations, with their peak values proportional to class probabilities; then the heatmaps of all $Q$ queries are stacked to form a single probability map. Note that the initial query embedding (the first column) produces an equivocal keypoint distribution. There exists confusion of keypoint locations in the first several layers of decoder, yet as the decoder layer goes deeper, the refinement proceeds and eventually yields a salient keypoint probability map (the last column).

\section{Conclusion}
In this paper, we have presented Pose Regression TRansformer (PRTR), a new design for regression-based multi-person pose recognition method based on the Transformer structure \cite{vaswani2017attention,carion2020end}. It treats the pose recognition task as a regression task, removes complex pre/post-processing procedures and requires fewer heuristic designs compared with existing heatmap-based approaches. Our method includes two alternatives, one as a two-stage and the other an end-to-end one. PRTR achieves state-of-the-art performance compared with other existing regression-based methods on the challenging COCO dataset. Distribution and refinement visualization of keypoint queries blazes the trail of revealing Transformer decoder inner mechanisms. In the future, we would like to investigate more powerful backbone networks and combine regression-based human detection and pose recognition in a more flexible manner.

{\small
\bibliographystyle{ieee_fullname}
\bibliography{egbib}
}
\clearpage

\end{document}